\pdfoutput=1

\documentclass[11pt]{article}

\usepackage[]{acl}

\usepackage{times}
\usepackage{latexsym}
\usepackage{comment}

\usepackage[T1]{fontenc}

\usepackage[utf8]{inputenc}

\usepackage{microtype}

\title{IISERB Brains at SemEval 2022 Task 6: A Deep-learning Framework to Identify Intended Sarcasm in English}
\usepackage{authblk}

\author{Tanuj Singh Shekhawat$^{*1}$, Manoj Kumar$^{*2}$, Udaybhan Rathore$^{*3}$, Aditya Joshi$^4$, Jasabanta Patro$^5$ \\ 
$^{1235}$ Indian Institute of Science Education and Research, Bhopal \\ $^4$ Seek.com.au\\ \{$^1$tanuj19, $^2$manoj19, $^3$udaybhan19, $^5$jpatro\}@iiserb.ac.in, $^4$ajoshi@seek.com.au}

\begin{document}
\maketitle 
\begin{abstract}

This paper describes the system architectures and the models submitted by our team "IISERB Brains" to SemEval 2022 Task 6 competition. We contested for all three sub-tasks floated for the English dataset. On the leader-board, we got \textbf{19th rank} out of \textbf{43 teams} for sub-task A, \textbf{8th rank} out of \textbf{22 teams} for sub-task B, and \textbf{13th rank} out of 16 teams for sub-task C. Apart from the submitted results and models, we also report the other models and results that we obtained through our experiments after organizers published the gold labels of their evaluation data. All of our code and links to additional resources are present in GitHub\footnote{\url{https://github.com/manojmahan/iSarcasmEval-Intended-Sarcasm-Detection-In-English-main}} for reproducibility. \textit{Authors with equal contributions are marked by *}.

\end{abstract}

\section{Introduction}

Sarcasm in spoken or written form refers to a verbal irony that indicates the difference between the literal and intended meanings of an utterance \cite{joshi2017automatic}. While many people use it as a bitter remark to mock or ridicule a target\cite{patro-etal-2019-deep}, some also use it as a joke to amuse others\cite{joshi2015harnessing}. Sarcasm often used together or interchangeably with other ironic categories, is considered an essential component of human communication. A large portion of the web and social media text is sarcastic, which creates a challenge for traditional natural language processing (NLP) systems on basic tasks like sentiment classification, opinion mining, harassment detection, author profiling etc. Since these systems are deployed widely across various industries, administration, data analysts etc, designing a robust sarcasm detecting component would help the downstream tasks substantially. The SemEval 2022 task 6\cite{abufarha-etal-2022-semeval} identifies some of the challenges persisting till now, particularly in English and Arabic texts. Our team participated in the tasks floated for the English language under the name "IISERB Brains". The organizers have floated three sub-tasks: (i)Sub-task A to detect whether a given text is sarcastic, (ii)Sub-task B to identify which ironic category the sarcastic tweet belongs to, and (iii) Sub-task C to given two pieces of text, identify the sarcastic one.  

In this paper, we report, not only the details of our participating systems and the results we got for them on the evaluation data released by the organizers but also additional experiments we did and results we got for them after the evaluation period. Particularly, section \ref{sec:relatedworks} highlights the literature on and around the sub-tasks, section \ref{sec:background} illustrates the dataset and task details, section \ref{sec:sysoverview} reports the details of our systems followed by section \ref{sec:expandres} which describes the experiments and results. Finally, in section \ref{sec:conclusion} we conclude our work by reiterating our overall results.

With our submitted systems, we got \textbf{19th rank} out of \textbf{43 teams} for sub-task A, \textbf{8th rank} out of \textbf{22 teams} for sub-task B, and \textbf{13th rank} out of 16 teams for sub-task C. Our code is made public to ensure the reproducibility of our results.

\section{RelatedWorks}
\label{sec:relatedworks}

Sarcasm has been an interesting research topic for computational linguists for a long time\cite{joshi2017automatic, pilling2017human}. Researchers have studied the topic from depth(sarcasm sub-categories, linguistic nuances etc.) and breadth(multi-modal sarcasm\cite{castro2019towards}, multi-lingual sarcasm \cite{liu2021does, bansalcode},  sarcasm targets \cite{patro-etal-2019-deep} etc). A considerable amount of work has been reported on the task 'sarcasm detection' itself. This task particularly deals with identifying whether a given text is sarcastic. 
The sub-task A proposed by the organizers also belongs to this task. \\\\
While the traditional approaches have used feature-based machine learning models, recent approaches mostly relied on deep-neural models to report state of art performance. Lexical features like emoticons, special characters, word-patterns have always been preferred features along with semantic features like parts-of-speech tags, contrasting sentiments etc\cite{davidov2010enhanced}\cite{veale2010detecting} \cite{riloff2013sarcasm}\cite{joshi2015likert}\cite{ghosh2018sarcasm} in traditional approaches. Deep-neural methods, on the other hand, learn latent features for the same task. They rely on architectures like RNNs, CNNs, transformers etc. In their models \cite{joshi2017automatic, pilling2017human, tay2018reasoning, tarunesh2021trusting, jaiswal2020neural}. Pre-trained language models like BERT, RoBERTa, XLNet etc, have been extensively used as token-encoders in these models.    

Identifying sarcastic intention has always been a challenging task, even humans sometimes have difficulties. Recently, researchers have started focusing on contextual information such as author context \cite{ghosh2020report}, multi-modal context \cite{ghosh2020report}, eye-tracking information \cite{govindan2018big}, or conversation context \cite{ghosh2020report, srivastava2020novel} to capture it. Related to shared tasks on figurative language analysis, recently, \citet{van2018automatic} has conducted a SemEval task on irony detection in Twitter focusing on the utterances in isolation. 

\section{Background:}
\label{sec:background}

\subsection{Dataset details:}

We used the English dataset provided by the task organizers\cite{abufarha-etal-2022-semeval}. It has 3468 English tweets. It has 867 sarcastic and 2601 non-sarcastic tweets, which indicates that the dataset is highly unbalanced. For each sarcastic tweet, the organizers have also provided the ironic sub-classes to which the tweet belongs. The sub-classes are sarcasm, irony, satire, understatement, overstatement, and rhetorical question. Also, there are many tweets with multiple labels assigned to them. The distribution of tweets over their labels are shown in Table \ref{tab:InputDataDist}. 

According to the organizers, unlike in the other publicly available datasets, there in literature for the same purpose. In this dataset, the sarcastic label of each tweet is marked by its author. This makes the dataset unique in terms of capturing the sarcastic intention of the tweet-author(s). Additionally, for each sarcastic tweet, the organizers have asked the its authors to rephrase the tweet-text to convey the same message without using sarcasm. However, the organizers have relied on linguistic experts to annotate the sub-categories. Experts referred \citet{leggitt2000emotional} for the definitions of sub-categories.


\begin{table}
\centering
\begin{tabular}{|p{5cm}|p{0.7cm}|} \hline
 \textbf{Labels} & \textbf{|N|}  \\ \hline
 Non-sarcastic & 2601  \\
 Only-sarcasm & 568 \\
 Only-irony & 122 \\
 Sarcasm and irony & 1 \\
 Sarcasm and satire & 21  \\
 Sarcasm and overstatement &  31 \\
 Sarcasm and understatement & 6 \\
 Sarcasm and rhetorical questions & 86 \\
 Irony and satire & 4\\
 Irony and overstatement & 9\\
 Irony and understatement & 4\\
 Irony and rhetorical question & 15\\
Understatement and rhetorical question & 2\\
Irony, understatement and rhetorical question & 1 \\
Sarcasm, understatement and rhetorical question & 1\\ \hline
 \textit{Total} & 3468\\ 
 \hline
\end{tabular}
\caption{ Label-wise distribution of tweets }
    \label{tab:InputDataDist}
\end{table}

\subsection{Task details:}

Based on the dataset they have released, the task organizers have formulated three challenges as sub-tasks. The details of the sub-tasks are,

\begin{itemize}
    \item \textbf{Sub-task A:} Given a text, determine whether it is sarcastic or non-sarcastic;
    \item \textbf{Sub-task B:} This sub-task is designed for particularly English dataset. It is a binary multi-label classification task. Here, given a text, we have to determine which ironic speech category it belongs to, if any;
    \item \textbf{Sub-task C:} Given a sarcastic text and its non-sarcastic rephrase, i.e. two texts that convey the  same  meaning,  determine  which  of the two is the sarcastic.
\end{itemize}

For all of the three sub-tasks, the organizers have informed us that precision, recall, accuracy, and macro-F1 of the participating models will be reported. According to them the main metrics of evaluation for the sub-task A is the F1-score for the sarcastic class. Similarly, for sub-task B and sub-task C, it is the macro-F1 score and the accuracy,  respectively.  

\section{System overview:}
\label{sec:sysoverview}

\subsection{Additional resources:}
\label{sec:addiresou}

As shown in Table \ref{tab:InputDataDist}, the dataset is highly unbalanced, and the sample size is small. To mitigate this issue, we considered additional publicly available datasets published earlier for a similar task with some synthetically generated text. The details of such datasets are following,

\begin{itemize}
    \item \textbf{SemEval 2018 task 3:} We used the training and test data provided by SemEval 2018 task on irony detection \cite{van2018semeval}\footnote{\url{https://github.com/zeroix15/Twitter_Sarcasm_Detections/tree/main/semeval18}} as an additional resource for training of our models.
    
    \item \textbf{MUStARD:} We used the textual part of the multi-modal sarcasm detection dataset provided by \citet{castro-etal-2019-towards}\footnote{\url{https://github.com/soujanyaporia/MUStARD/blob/master/data/sarcasm_data.json}} as an additional resource for training of our models.
    
    \item \textbf{FigLang 2020 Sarcasm:} We used the sarcasm dataset\footnote{\url{https://github.com/EducationalTestingService/sarcasm/releases}} released as a part of shared task of FigLang2020 workshop \footnote{\url{https://sites.google.com/view/figlang2020/home}} as additional data for training of our models. 
    
    \textbf{Augmentation:} For increasing the instances labeled with sub-categories in the train data provided by the task organizers, we performed data augmentation using the python nlpaug\footnote{\url{https://nlpaug.readthedocs.io/en/latest/overview/overview.html}} library. We took sarcastic tweets given by organizers used the word-replacement procedure provided by nlpaug to synthesize three additional tweets from each input tweet. We used 'distilbert-base-uncased' \footnote{\url{https://huggingface.co/distilbert-base-uncased}} contextual embedding as the input embedding for this process. 
\end{itemize}

The statistical distribution of additional resources is shown in Table \ref{tab:data_source}. After elimination of the duplicates, the final dataset had 19986 tweets. 


\begin{table}
    \centering
    \begin{tabular}{|c|c|}
    \hline
    Source & |Instances|  \\
    \hline
    SemEval22 Task 6  &  3468 \\
    SemEval18 Task 3 training  &  3398 \\
    SemEval18 Task 3 test  &  780 \\
    MUStARD & 690 \\
    FigLang20 Sarcasm  & 9400  \\
    Data Augmentation (867x3) & 2601 \\
    \hline
    \end{tabular}
    \caption{Basic statistics of additional sources. }
    \label{tab:data_source}
\end{table}

\subsection{Data preprocessing}
\label{SubSec:Preprocess}

We followed the following preprocesing steps for every instance in our dataset. 

\begin{itemize}
    
    \item \textbf{Case conversion:} We converted the dataset into lowercase except for those words in which the whole word is in uppercase.
    
    \item \textbf{Stop-word removal:} We removed all stop-words as they contain low information. We did this using python NLTK library\footnote{\url{https://www.nltk.org/}}.
    
    \item \textbf{Data cleaning:} We did basic data cleaning which include removal of links, punctuation marks, floating point(.) characters and username. However, we didn't apply stemming and lemmatization techniques because we believe they will distort the meaning of instances.
    
    \item \textbf{Special Tokens:} We added special tokens at the starting and ending of the instances as required by different tokenizers for respective transformer based models. 
    
\end{itemize}
    
\subsection{Model description:}
\label{sec:modeldescription}

We relied on transformer based architectures to design our models for all sub-tasks. This is because dominate as state of the art in almost all NLP tasks. We built our models using the huggingface transformer library\footnote{\url{https://huggingface.co/docs/transformers/index}}. They support generic transformer based architectures with the ability to seamlessly initialize the tokens with different pre-trained embeddings. 

\begin{itemize}
    \item \textbf{Sub-task A:} For this sub-task, we deployed the binary classifier versions of different transformer based architectures provided by the huggingface transformer library. We particularly experimented with BERT\cite{devlin-etal-2019-bert}, RoBERTa\cite{liu2019roberta}, XLNet\cite{yang2019xlnet} and DistilBERT\cite{sanh2019distilbert} architectures. Apart from initializing the tokens with respective pre-trained embeddings, we fine-tuned the last layers of the models according to our training data. Further, we added the non-sarcastic versions of 867 sarcastic tweets given by the organizers to the training set.        
    \item \textbf{Sub-task B:} Here, instead of using a multi-label classifier, we used six binary classifier versions of the transformer based architectures provided by the same huggingface library. Like in the previous sub-task, here we have also experimented with BERT, RoBERTa, XLNet, and DistilBERT architectures. We constructed training data( refer Table \ref{tab:data_source_taskB}) for label-wise models to fine-tune it. We merged the predictions of all six models to get the final prediction labels. 
    
    \item \textbf{Sub-task C:} We formulated this sub-task as a parallel combination of two sub-task A models. We considered the same architecture for both parallel sub-components in all of our experiments. 
    
\end{itemize}

\section{Experiments and results}
\label{sec:expandres}

\subsection{Experiments and results for sub-task A:} For this sub-task, in addition to the data given by the organizers, we considered the other datasets as mentioned in Table \ref{tab:data_source}. We applied the pre-processing(details in section \ref{SubSec:Preprocess}) steps before sending them to the respective tokenizers of the considered architectures. The tokens are initialized by respective pre-trained embeddings. The dataset is divided into three parts i.e. training, validation and test set with a ratio of 0.7:0.2:0.1 for the parameter and hyper-parameter tuning. We generated the predictions for the unlabeled data provided by the organizers and submitted them in the codalab submission site\footnote{\url{https://codalab.lisn.upsaclay.fr/competitions/1340}} for evaluation. As the organizers have later released the labels for their evaluation data, we can compare all our models by ourselves too. The performance of our models for the evaluation data is reported in Table \ref{tab:sub-task-a-results}. The number of epochs for which the models are trained are different for different models. We trained until our models started over-fitting. Note that the best performing result reported in the table is different from that we have submitted in the codalab site. We experimented with our models even after the evaluation period and the results in Table \ref{tab:sub-task-a-results} show the best performance we have achieved till date. The models submitted as a part of competition is reported in Table \ref{tab:OurSubmission}. The hyper-parameters for all models are reported in Table \ref{tab:hyperparameters}.

\begin{table*}
    \centering
    \begin{tabular}{|c|c|c|c|c|c|}
    \hline
    Model  & Precision & Recall & F-1 score & F-1 sarcastic & accuracy \\
    \hline
    DistilBERT & 0.57& 0.64 & 0.53 & 0.34 & 0.62 \\
    \hline
    XlNet & 0.63& 0.61 & 0.62 & 0.34 & 0.82 \\
    \hline
    BERT & 0.60& 0.69 & 0.60 & 0.39 & 0.70 \\
    \hline
    \textbf{RoBERTa} & 0.70 & 0.68 & 0.69 & 0.47 & 0.86 \\
    \hline
    \end{tabular}
    \caption{Performance measures of our models submitted for sub-task A} 
    \label{tab:sub-task-a-results}
\end{table*}

\begin{table*}
    \centering
    \begin{tabular}{|c|c|c|c|c|c|}
    \hline
    Model  & Learning-rate & MAX\_SEQ\_LEN & BATCH\_SIZE & EPOCHS \\
    \hline
    RoBERTa & 2e-6 & 256 & 16 & 10\\
    \hline
    BERT & 2e-5 & 128 & 32 & 3\\
    \hline
    Xlnet & 2e-5 & 128 & 32 & 4\\
    \hline
   DistilBERT & 5e-5 & 1213 & 16 & 5\\
    \hline
    \end{tabular}
    \caption{Hyper-parameters of our models} 
    \label{tab:hyperparameters}
\end{table*}

\begin{table}
    \centering
    \begin{tabular}{|c|c|}
    \hline
    \multicolumn{2}{|c|}{SubTask A}\\
    \hline
    Model & F1 \\
    
    BERT & 0.34 \\
    \hline
    \multicolumn{2}{|c|}{SubTask B}\\
    \hline
    Model & Macro-F1\\
    
    BERT & 0.0751\\
    \hline
    \multicolumn{2}{|c|}{SubTask C}\\
    \hline
    Model & Accuracy\\
    
    BERT & 0.62\\
    \hline
    \end{tabular}
    \caption{Our submitted models and results for which we got ranks in three sub-tasks.} 
    \label{tab:OurSubmission}
\end{table}

\subsection{Experiments and results for sub-task B:} As stated in the previous section, for this sub-task we considered separate binary classifiers for each label. The sample size of the datasets created for individual classifiers are shown in Table \ref{tab:data_source_taskB}. Note that we didn't include the non-sarcastic tweets provided by the organizers in our new datasets. Rather, we added the synthetic tweets generated by the nlpaug library (see section \ref{sec:addiresou}) to amplify the label for which it is created. We didn't generate any additional synthetic text for the dataset corresponds to 'sarcasm' label as it is the dominant class in the provided 'sarcastic' data. Thus, the dataset created for the binary classification of 'sarcasm' class has the original 867 sarcasic tweets. In other label-specific datasets we increased the corresponding label tweets with the help of nlpaug library. As stated in section \ref{sec:addiresou}, we did this by generating three similar texts for each tweet tagged with the considered label. Thus, the newly created label-specific datasets have different sample size as shown in Table \ref{tab:data_source_taskB}.  We fine-tuned BERT model for each category of ironic speech. The organizers have evaluated sub-task B based on macro-F1 score. The results as reported in Table \ref{tab:my_label1}. 

\begin{table*}
    \centering
    \begin{tabular}{|c|c|c|c|c|c|c|}
    \hline
    Source & sarcasm & irony & overstatement &  rhetorical question &  satire & understatement \\
    \hline
    SemEval22 Task 6  &  867 & 867 & 867 & 867 & 867 & 867 \\
    Data Augmentation &0&465&120&303&75&30 \\ 
    \hline
    Total & 867 & 1332 & 906 & 1170 & 942 & 98\\
    \hline
    \end{tabular}
    \caption{Datasets sources for sub-task B}
    \label{tab:data_source_taskB}
\end{table*}

\begin{table}
    \centering
    \begin{tabular}{|c|c|c|}
    \hline
    Results  & BERT  \\
    \hline
    Macro F-1 & 0.0751  \\
    \hline
    F1-Sarcasm & 0.2294 \\
    \hline
    F1-irony & 0.0963  \\
    \hline
    F1-satire & 0.0833  \\ 
    \hline
    F1-understatement & 0.0000  \\  
    \hline
    F1-overstatement & 0.0000  \\
    \hline
    F1-rhetorical question & 0.0414  \\
    \hline
    \end{tabular}
    \caption{The macro-f1 and class-wise f1 scores for sub-task B} 
    \label{tab:my_label1}
\end{table}

\textbf{Experiments and results for sub-task C}
As reported in section \ref{sec:modeldescription}, we formulated the task as two parallel combination of sub-task A models. The considered same model across the parallel sub-components. The accuracy(evaluating measure considered by the organizers) of different models are reported in Table \ref{tab:my_label2}. As we can infer, BERT based fine tuned model performed best on the evaluation data among all with an accuracy of \textbf{0.62}.

\begin{table}
    \centering
    \begin{tabular}{|c|c|c|c|c|c|c|}
    \hline
    Model  & accuracy  \\
    \hline
    RoBERTa & 0.47 \\
    \hline
    XlNet & 0.49\\
    \hline
    DistilBERT & 0.57\\
    \hline
    BERT & 0.62 \\
    \hline
    \end{tabular}
    \caption{Accuracy of our models on evaluation data provided for sub-task c.} 
    \label{tab:my_label2}
\end{table}

\section{Conclusion}
\label{sec:conclusion}

In this paper, we discussed our models for different sub-tasks proposed by SemEval 2022 task 6 organizers for the English dataset. We have experimented with well-known transformer architectures with different pre-trained language models.

With our submitted models, our team "IISERB Brains", got \textbf{19th rank} out of \textbf{43 teams} on sub-task A, \textbf{8th rank} out of \textbf{22 teams} on sub-task B and \textbf{13th rank} out of 16 teams in sub-task C. All of our code and links to considered data are uploaded in GitHub\footnote{\url{https://github.com/manojmahan/iSarcasmEval-Intended-Sarcasm-Detection-In-English-main}} for reproduciblity.

\bibliography{main}

\end{document}